\documentclass{article}
\usepackage{ijcai24}

\usepackage{times}
\usepackage{soul}
\usepackage{url}
\usepackage[hidelinks]{hyperref}
\usepackage[utf8]{inputenc}
\usepackage[small]{caption}
\usepackage{graphicx}
\usepackage{amsmath}
\usepackage{amsthm}
\usepackage{booktabs}
\usepackage{algorithm}
\usepackage{algorithmic}
\usepackage[switch]{lineno}

\usepackage{booktabs, colortbl}
\usepackage{multirow}
\usepackage[table]{xcolor}
\usepackage{graphicx} 

\title{AesBench: An Expert Benchmark for Multimodal Large Language Models on Image Aesthetics Perception}

\author{
Yipo Huang$^{1 \dagger}$\and
Quan Yuan$^{1 \dagger}$\and
Xiangfei Sheng$^1$\and
Zhichao Yang$^1$\and
Haoning Wu$^2$\and
Pengfei Chen$^1$\and
Yuzhe Yang$^3$\and
Leida Li$^{1 *}$\And
Weisi Lin$^2$
\affiliations
$^1$Xidian University $^2$Nanyang Technological University $^3$OPPO Research Institute
\emails
\{huangyipo, dylan.yuanquan, xiangfeisheng, yangzhichao\}@stu.xidian.edu.cn,
haoning001@e.ntu.edu.sg, ippllewis@gmail.com,
\{chenpengfei, ldli\}@xidian.edu.cn,
wslin@ntu.edu.sg\\
$^\dagger$Equal contribution. $^*$Corresponding author.
}

\begin{document}

\maketitle

\begin{abstract}
With collective endeavors, multimodal large language models (MLLMs) are undergoing a flourishing development. However, their performances on image aesthetics perception remain indeterminate, which is highly desired in real-world applications. An obvious obstacle lies in the absence of a specific benchmark to evaluate the effectiveness of MLLMs on aesthetic perception. This \textbf{\emph{blind}} groping may impede the further development of more advanced MLLMs with aesthetic perception capacity. To address this dilemma, we propose \textbf{AesBench}, an expert benchmark aiming to comprehensively evaluate the aesthetic perception capacities of MLLMs through elaborate design across dual facets. (1) We construct an Expert-labeled Aesthetics Perception Database (EAPD), which features diversified image contents and high-quality annotations provided by professional aesthetic experts. (2) We propose a set of integrative criteria to measure the aesthetic perception abilities of MLLMs from four perspectives, including \textbf{Perception} (\emph{AesP}), \textbf{Empathy} (\emph{AesE}), \textbf{Assessment} (\emph{AesA}) and \textbf{Interpretation} (\emph{AesI}). Extensive experimental results underscore that the current MLLMs only possess rudimentary aesthetic perception ability, and there is still a significant gap between MLLMs and humans. We hope this work can inspire the community to engage in deeper explorations on the aesthetic potentials of MLLMs. Source data will be available at \url{https://github.com/yipoh/AesBench}.
\end{abstract}

\section{Introduction}
Recent advancements in large language models (LLMs), such as ChatGPT \cite{chatgpt} and LLaMA \cite{LLaMA}, have demonstrated remarkable achievements in comprehension, reasoning, and generation, garnering substantial interest from both academia and industry. Building on the success of LLMs in pure-textual tasks, multimodal large language models (MLLMs) have sparked a new wave in the field of visual-language processing, releasing numerous representative works, such as MiniGPT \cite{MiniGPT-4}, Otter \cite{Otter} and LLaVA \cite{LLAVA}. Recently, these MLLMs have evolved into versatile assistants, facilitating human-machine interaction and collaboration \cite{Q-Boost}. Despite the broad capabilities of MLLMs, existing benchmarks mainly focus on general evaluation in several language or vision tasks, \emph{e.g.} visual question answering \cite{VQA}, image captioning \cite{Captioning}, object segmentation \cite{Segmentation} and content understanding \cite{BenchLMM}. However, \textbf{the effectiveness of MLLMs on the highly abstract image aesthetics perception task remains underexplored}, which plays a significant role in image aesthetics assessment, aesthetic attribute analysis, image aesthetics cropping and image aesthetics captioning \cite{SARQUE}. The potential of MLLMs in image aesthetics perception holds considerable promise for applications in smart photography, album management, photo recommendation and image enhancement, \emph{etc} \cite{MM3,DSS}. Therefore, it is highly desired to evaluate the abilities of existing MLLMs on aesthetic perception, which is expected to promote the further development of more advanced MLLMs with better aesthetic perception.

  \begin{figure*}[!t]
    \centerline{\includegraphics[width=17.5cm]{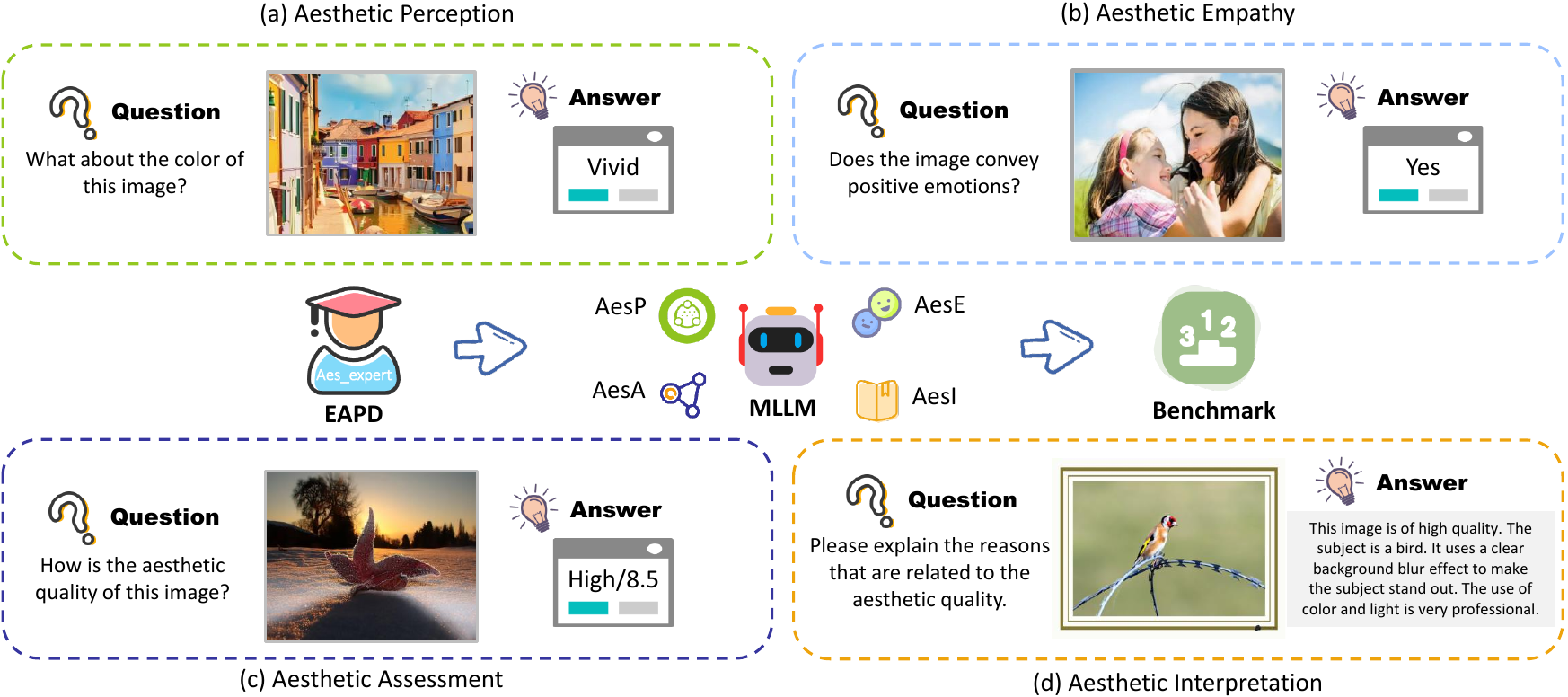}}
    \caption{Overview of the proposed AesBench. We first construct a high-quality Expert-labeled Aesthetics Perception Database (EAPD), based on which we further build the \emph{golden} benchmark to evaluate four abilities of MLLMs on image aesthetics perception, including Aesthetic Perception (AesP), Aesthetic Empathy (AesE), Aesthetic Assessment (AesA) and Aesthetic Interpretation (AesI).}
    \label{expert}
    \end{figure*}

In this work, we propose \textbf{AesBench}, a systematically designed \emph{golden} benchmark to comprehensively evaluate the aesthetic perception abilities of MLLMs. Specifically, the proposed AesBench encompasses two major facets, \emph{i.e.} \textbf{high-quality data} and \textbf{integrative criteria}. In terms of the former, we construct the Expert-labeled Aesthetics Perception Database (EAPD), which consists of 2,800 diverse-sourced images covering natural images (NIs), artistic images (AIs) and artificial intelligence-generated images (AGIs). Furthermore, each image of the EAPD is annotated by experts with professional backgrounds in aesthetics, which comprises researchers engaged in computational aesthetics, educators versed in aesthetic principles, and art students possessing sophisticated art skills. With these high-quality annotations,  the benchmark results of MLLMs can be more reliable. For the latter, we develop a set of integrative criteria to systematically evaluate the aesthetic perception abilities of MLLMs. These criteria are 
designed from four dimensions, which are illustrated in Figure \ref{expert}. 

  \begin{itemize}
    \item \textbf{Aesthetic Perception} (\emph{AesP}): As shown in Figure \ref{expert}(a), this dimension focuses on the ability of MLLMs to recognize and understand aesthetic attributes.
    \item \textbf{Aesthetic Empathy} (\emph{AesE}): As shown in Figure \ref{expert}(b), this dimension evaluates the ability of MLLMs to resonate with the emotional aspects conveyed through aesthetic expressions like humans.
    \item \textbf{Aesthetic Assessment} (\emph{AesA}): As shown in Figure \ref{expert}(c), this dimension evaluates the ability of MLLMs to judge aesthetic grades and predict quality scores based on the language description.
    \item \textbf{Aesthetic Interpretation} (\emph{AesI}): As shown in Figure \ref{expert}(d), this dimension involves the ability of MLLMs to interpret and analyze the reasons for aesthetic quality.
  \end{itemize}

In summary, we present a comprehensive exploration of the abilities of MLLMs for the task of image aesthetics perception, which is structured around a novel four-dimensional evaluation framework, encompassing perception, empathy, assessment, and interpretation. These dimensions collectively form the foundation of AesBench, a \emph{golden} benchmark of MLLMs on image aesthetics perception. The contributions of this work can be summarized as three-fold.

$\bullet$  We construct a high-quality Expert-labeled Aesthetics Perception Database (EAPD), encompassing 2,800 diverse-sourced images, along with 8,400 question-answer pairs and 2,800 aesthetic interpretations, all meticulously annotated by professional experts in aesthetics. 

$\bullet$  We propose a set of integrative criteria to systematically evaluate the aesthetic perception abilities of MLLMs, which is based on the four-dimensional evaluation framework, including perception (AesP), empathy (AesE), assessment (AesA), and interpretation (AesI). 

$\bullet$  We conduct extensive evaluations on 15 well-known MLLMs using AesBench, including two authoritative GPT-4V and Gemini Pro Vision, as well as thirteen state-of-the-art open-source counterparts. Comprehensive experimental results reveal that there is still a significant gap between MLLMs and humans in image aesthetics perception.

We will make our code and database publicly available. We hope that these contributions can encourage the community to delve into more profound investigations of the yet untapped aesthetic potentials of MLLMs in future studies.



\section{Constructing the EAPD}



Image aesthetics perception is widely regarded as a highly abstract and complex task \cite{AesCLIP,PARA,TAVAR}. To ensure the reliability of AesBench, it is essential to construct a high-quality Expert-labeled Aesthetics Perception Database (EAPD), meticulously annotated by professional experts in aesthetics. Specifically, the construction of EAPD consists of three stages, including data collection, expert selection and subjective experiments.

\subsection{Data Collection}

To cover diversified image contents, we collect multi-sourced images from diverse datasets, including three \emph{natural-image} datasets (\emph{i.e.} AADB \cite{AADB}, PARA \cite{PARA} and TAD66K \cite{TAD66K}), three \emph{artistic image} datasets (\emph{i.e.} BAID \cite{BAID}, CAD \cite{CAD} and ArtEmis \cite{ArtEmis}), as well as three \emph{artificial intelligence-generated image} datasets (\emph{i.e.} DiffusionDB \cite{DiffusionDB}, AGIQA-1K \cite{AGIQA-3K} and AGIQA-3K \cite{AGIQA-1K}). Then, we use a well-trained scene classification model \cite{CLIP} to predict the scene labels automatically, and then double-check the labels manually to maintain annotation correctness. Next, we sample 2,493 images based on scene labels to maintain content diversity. Further, we add 152 high-aesthetic images from LITE \cite{LITE} and Impressions \cite{Impressions} datasets, and 155 low-aesthetic images from SPAQ \cite{SPAQ} and KonIQ-10K \cite{KonIQ-10K} datasets, to balance aesthetic quality distribution. The information of sampled multi-sourced images is summarized in Table \ref{datasets}. Finally, the EAPD consists of \textbf{2,800 multi-sourced images}, which are provided to professional aesthetic experts for high-quality annotation. 

\subsection{Expert Selection}

\begin{table}[!t]
\caption{Overview of the image source datasets.}  \label{datasets}
\renewcommand{\arraystretch}{1.5}
\fontsize{8.5pt}{7pt}\selectfont %
\centering
\setlength{\tabcolsep}{1.9mm}{
\begin{tabular}{c|lc}
\toprule
Type            &Dataset &Sampled Size \\
      \midrule 
 \multirow{7}*{NIs}               
 &AADB \cite{AADB} &96  \\
 &PARA \cite{PARA} &539  \\
 &TAD66K \cite{TAD66K} &304    \\
 &LITE \cite{LITE}&146  \\
 &Impressions \cite{Impressions} &6    \\
 &SPAQ \cite{SPAQ} &50   \\
 &KonIQ-10K \cite{KonIQ-10K} &105   \\

\midrule 

  \multirow{3}*{AIs}             
  &BAID \cite{BAID}&622   \\
  &CAD \cite{CAD}&10   \\
  &ArtEmis \cite{ArtEmis}&170  \\
  \midrule 
 \multirow{3}*{AGIs}
  &DiffusionDB \cite{DiffusionDB}  &653   \\
 &AGIQA-3K \cite{AGIQA-1K}  &82   \\ 
 &AGIQA-1K \cite{AGIQA-3K} &17  \\

\bottomrule
\end{tabular}}
\end{table}

Unlike most existing aesthetics-related datasets, \textbf{all participants in this work are expected to be experts with rich aesthetic experience}. Specifically, to guarantee quality and diversity of annotation, we recruit subjects by considering six perspectives, including \emph{health state}, \emph{age distribution}, \emph{gender balance}, \emph{educational background}, \emph{working experience}, and \emph{personality traits}. First, we ensure that each participant is in \textbf{good health}, strictly adhering to ethical guidelines that prohibit any form of coerced participation. Second, we require that all participants should have \textbf{professional backgrounds in aesthetics}, which include researchers engaged in computational aesthetics, educators versed in aesthetic principles, and art students possessing sophisticated art skills. Third, we pay great attention to ensure \textbf{diversity among all participants}, particularly in \emph{age distribution}, \emph{gender balance}, \emph{good educational background}, and \emph{personality traits} (based on Big-Five personality traits\cite{PARA}). Taking into account the above factors comprehensively, we recruited 32 professional aesthetic experts who met the aforementioned criteria to participate in the subjective experiment, ensuring the validity and reliability of annotations. The detailed portrait distribution of participants is provided in {\color{gray}Supplementary Section 1}.

\subsection{Subjective Experiments}

\begin{figure*}[!t]
    \centerline{\includegraphics[width=18cm]{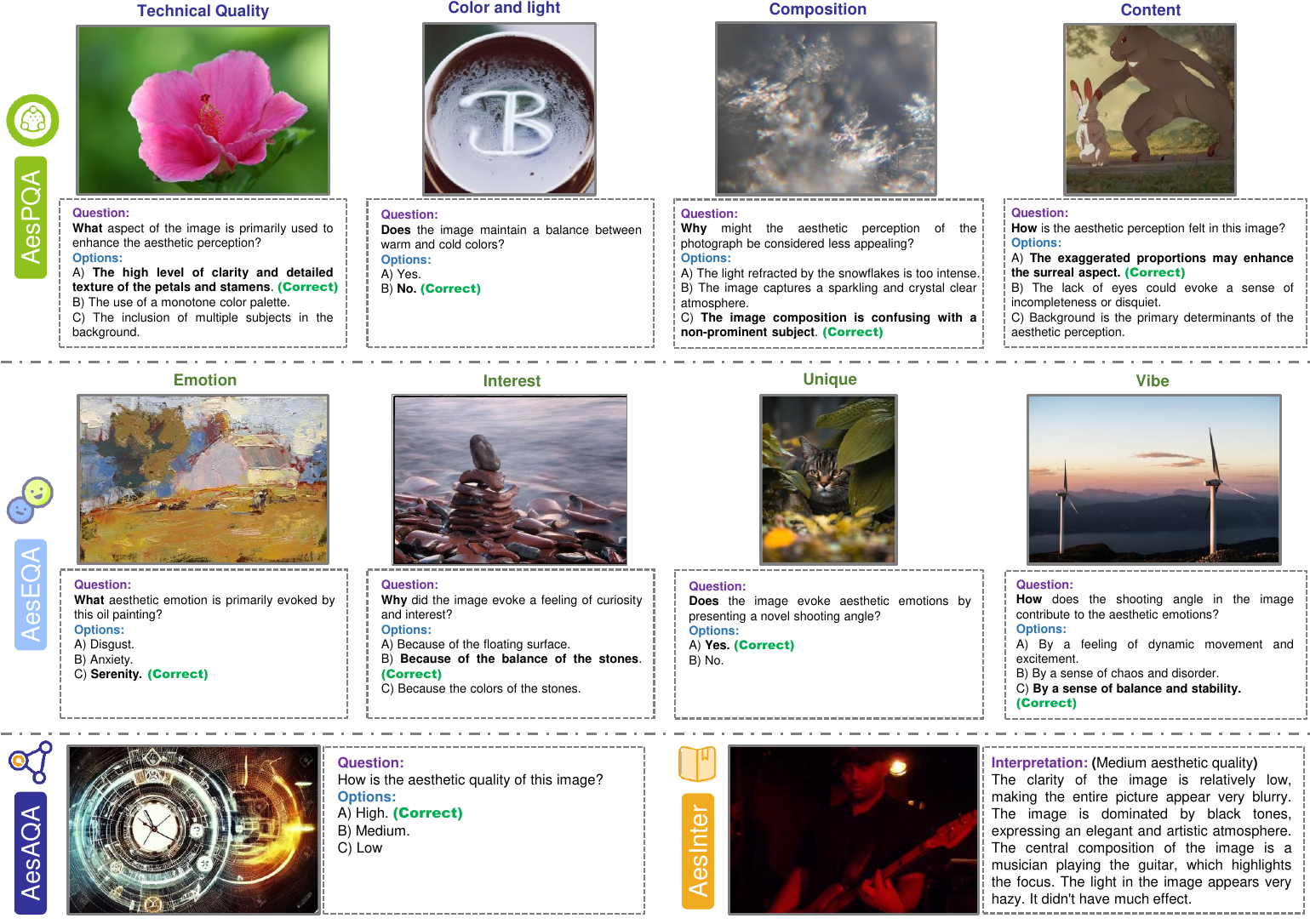}}
    \caption{Data samples from the constructed EAPD, which includes four subsets, \emph{i.e.} AesPQA subset (benchmarking AesP), AesEQA subset (benchmarking AesE), AesAQA subset (benchmarking AesA) and AesInter subset (benchmarking AesI). }
    \label{questions}
    \end{figure*}

 In this work, we conduct subjective experiments to collect expert annotations, in strict compliance with the generic psychological experiment protocol \cite{Evaluation}. First, the images of the entire database are segmented into 56 distinct sessions, and each session contains 50 unlabeled images. Then, we develop a local-deployed annotation tool (.exe format) and assign a personal user ID to each participant. Furthermore, a well-devised annotation quality control strategy ensures that all qualified annotations are automatically saved based on text richness and diversity, and each image is annotated by at least one expert (details are provided in {\color{gray}Supplementary Section 1}). To mitigate the risk of test fatigue, consecutive annotations on more than 30 images will trigger a reminder of the break. Similar to the existing MLLM benchmarks \cite{EgoPlan-Bench,BenchLMM}, we collect a total of \textbf{8,400 question-answer pairs} and \textbf{2,800 aesthetic interpretations}. Specifically, each image is equipped with three expert-asked questions focusing on AesP, AesE and AesA respectively, and one expert-answered aesthetic interpretation for AesI. These \emph{golden} annotations serve as ground truth for evaluating the abilities of MLLMs. Therefore, the EAPD can be divided into four subsets based on the benchmark objectives, as illustrated in Figure \ref{questions}, \emph{i.e.} AesPQA subset (evaluating AesP), AesEQA subset (evaluating AesE), AesAQA subset (evaluating AesA) and AesInter subset (evaluating AesI). These subsets are described in detail below.

\section{Developing the Criteria}
Based on the expert-labeled aesthetics perception database, we propose a set of integrative criteria to measure the aesthetic perception abilities of MLLMs from four perspectives, including \textbf{Perception} (\emph{AesP}), \textbf{Empathy} (\emph{AesE}), \textbf{Assessment} (\emph{AesA}) and \textbf{Interpretation} (\emph{AesI}). 

\subsection{Benchmark on Aesthetic Perception}

We first leverage the AesPQA subset to evaluate the aesthetic \textbf{perception} (AesP) ability of MLLMs, aiming to investigate whether they can accurately respond to simple natural language queries related to aesthetic attributes. The AesPQA subset contains 2,800 multi-sourced images (as listed in Table \ref{datasets}), and each image is equipped with one perception-related question, one correct option, and 1-3 false options (as illustrated in Figure \ref{questions}).  Specifically, the questions in AesPQA cover four perceptual dimensions and four question types. These diverse questions will be fed into MLLMs for evaluation, and then their outputs are further calculated as the rate of correct answers.

\emph{Perceptual Dimensions:} Inspired by recent research on aesthetic perception \cite{PARA} and human perception process on image aesthetics \cite{TAVAR}, the questions of the AesPQA subset consist of four dimensions as follows:
  \begin{itemize}
    \item \textbf{Technical quality} focuses on fundamental aspects such as sharpness, contrast, exposure, noise, etc, which are pivotal in influencing the aesthetic appeal of an image, forming the cornerstone of overall visual attractiveness. 
    \item \textbf{Color and light} are crucial for both the overall aesthetic expression of an image and the local emphasis on objects, which involve a detailed evaluation of color contrast, color harmony, color richness, and use of light.
    \item \textbf{Composition} is not merely the visual location of the target but a fundamental aspect of conveying meaning and aesthetic value in images, \emph{e.g.} symmetry, rule of thirds, and leading lines, playing a crucial role in structuring the visual narrative and enhancing the aesthetic appeal.
    \item \textbf{Content} focuses on the thematic essence, personal character, and the manner of story presentation in an image, which is crucial for evaluating elements such as originality, creativity, attraction, and enjoyment.
  \end{itemize}

\emph{Question Types:} In the AesPQA subset, inspired by the existing MLLM benchmarks \cite{EgoPlan-Bench,BenchLMM}, four commonly used question types are utilized to simulate human query formats when interacting with MLLMs. 1) \textbf{Yes-or-No} questions are binary queries requiring a simple positive or negative response (\emph{e.g., Does the image maintain a balance between warm and cold colors?}). 2) \textbf{What} questions are used to examine more comprehensive aesthetic perception. For example, \emph{What aspect of the image is primarily used to enhance the aesthetic perception?} 3) \textbf{How} questions are employed to evaluate more details about the production of aesthetic feeling (\emph{e.g., How is the aesthetic perception felt in this image?}). 4) \textbf{Why} questions aim to understand the rationale behind aesthetic perception, including the reasons for aesthetic judgment or the underlying principles of visual appeal (\emph{e.g., Why might the aesthetic perception of the photograph be considered less appealing?}). The diversified question types effectively simulate a wide range of human-like inquiries, facilitating the practicability of AesBench in evaluating the aesthetic perception abilities of MLLMs.
 
\emph{Evaluation Process:} Based on the AesPQA subset, we feed the perception-related questions to various MLLMs to measure their abilities on aesthetic perception. The input prompt to query MLLMs is exemplified as \emph{\#User: Does the image maintain a balance between warm and cold colors?} [Image\_token] \emph{Choose one from the following options: A) Yes B) No. You should output the correct option without explanation.} Despite the explicit prompt for MLLMs to merely provide the correct option, as observed by recent works \cite{Q-Bench,BenchLMM,Q-Align} and further validated by us, some MLLMs are unable to consistently generate outputs that align with \textbf{instructed format}. For the above \emph{Yes-or-No} question with the correct option: \emph{B) No}, different MLLMs may output ``B) No", ``B", ``No", or ``The image can not maintain a balance between warm and cold colors", which are all correct in fact. These irregular formats present significant challenges for the accurate assessment by existing traditional language metrics (such as BLEU-4, CIDEr). Inspired by \cite{LLAVA}, we propose, validate, and implement a GPT-assisted evaluation strategy that integrates decision rules with GPT. Specifically, outputs of MLLMs that align strictly with the candidate options are regarded as `\emph{explicit responses}', determined through direct comparison with the correct option. Otherwise, the outputs are regarded as `\emph{ambiguous responses}'. For these, the questions, correct answers and \emph{ambiguous responses} of MLLMs are submitted to GPT for evaluation (details are provided in {\color{gray}Supplementary Section 2}).

\subsection{Benchmark on Aesthetic Empathy}
Aesthetic empathy, a critical aspect of image aesthetics perception, embodies a multi-dimensional and comprehensive process, involving emotional response, interest, recognition of uniqueness, and understanding of the vibe in aesthetic experiences. Therefore, based on the proposed AesEQA subset, we further study the abilities of MLLMs to understand the aesthetic empathy contained in images as humans. Like AesPQA, the AesEQA subset comprises 2,800 images, each accompanied by one empathy-related question, one correct option, and 1-3 false options, as illustrated in Figure \ref{questions}. These questions are fed into MLLMs for evaluation, and the efficacy of these models is quantified based on the accuracy rate of their responses. Specifically, based on the principles of sentiment analysis research \cite{EmP}, AesEQA collects empathy-related questions from four dimensions as follows:

  \begin{itemize}
    \item \textbf{Emotion} refers to the complex emotional reactions that images evoke. It includes a range of emotional responses, such as joy, sadness, fear, and surprise, which are triggered by the image.
    \item \textbf{Interest} is defined as the degree of curiosity and attraction towards images, such as visual allure, emotional resonance, novelty, facts and culture.
    \item \textbf{Uniqueness} evaluates the originality and distinctiveness of photographic images or artworks, \emph{e.g.} novel concepts, unique stylistic approaches and innovative techniques.
    \item \textbf{Vibe} refers to the overarching emotional ambiance or thematic representation conveyed by images, which articulates a holistic sensation, characterized by harmony, liveliness, mystery, elegance, and others.
  \end{itemize}

Each dimension provides a unique perspective for evaluating the empathy-related abilities of current MLLMs. Additionally, the AesEQA subset also integrates a quartet of question types: \emph{Yes-or-No, What, How} and \emph{Why}. These four types of questions are leveraged to examine the ability to deal with common cases of human-machine interaction and collaboration, which play a pivotal role in the thorough evaluation of MLLMs. Finally, we also employ the GPT-assisted evaluation strategy to calculate the accuracy rate of the responses generated by MLLMs. 

\begin{table*}[!t]
\caption{Evaluation results on the \textbf{Aesthetic Perception} ability. We highlight the best results of each sub-category in bold and the second-best results in underlining.}  \label{AesP}
\renewcommand{\arraystretch}{1.5}
\fontsize{8.5pt}{7pt}\selectfont %
\centering
\setlength{\tabcolsep}{1mm}{
\begin{tabular}{l|cccc|ccc|cccc|ccc}
\toprule
\multirow{2}*{MLLM}    & \multicolumn{4}{c|}{\textbf{Perceptual Dimensions}}& \multicolumn{3}{c|}{\textbf{Image Sources}} 
& \multicolumn{4}{c|}{\textbf{Question Types}}   & {\multirow{2}*{\textbf{Overall}}}& {\multirow{2}*{\textbf{Rank}}}\\
\cmidrule(lr){2-5}\cmidrule(lr){6-8}\cmidrule(lr){9-12}
&\emph{Tec. Qua.} &\emph{Col. Lig.} &\emph{Comp.} &\emph{Content} &\emph{NIs} &\emph{AIs} &\emph{AGIs }  &\emph{Yes-No } &\emph{What} &\emph{How} &\emph{Why}  \\
\midrule 
\rowcolor{lightgray!25}
\emph{Random guess}     &34.82\% &31.02\% &30.53\% &37.20\%   &31.59\% &32.50\% &31.19\%    &50.00\% &27.35\% &27.32\% &25.00\%       &35.02\% &{/} \\
\midrule 
Q-Instruct              &\underline{66.03\%} &\underline{74.48\%} &\textbf{73.68\%} &\textbf{68.09\%}   &\textbf{76.48\%} &\underline{69.70\%} &\underline{69.28\%}    &64.68\% &63.31\% &\textbf{85.28\%} &\underline{86.34\%}      &\textbf{72.61\%} &\textbf{1}\\
GPT-4V                  &\textbf{69.02\%} &\textbf{74.66\%} &71.72\% &65.57\%   &\underline{75.67\%} &\textbf{72.58\%} &65.82\%    &\underline{68.93\%} &\underline{64.67\%} &76.70\% &84.46\%       &\underline{72.08\%} &\underline{2}\\
Gemini Pro Vision       &65.08\% &74.57\% &\underline{72.24\%} &\underline{67.97\%}   &74.63\% &69.62\% &\textbf{70.03\%}    &64.70\% &\textbf{64.95\%} &\underline{78.71\%} &\textbf{90.24\%}       &71.99\% &3\\
ShareGPT4V              &62.18\% &71.90\% &69.29\% &64.89\%   &70.79\% &71.57\% &63.96\%    &\textbf{69.32\%} &61.33\% &72.01\% &77.56\%       &69.18\% &4\\
mPLUG-Owl2              &60.90\% &70.57\% &68.30\% &62.77\%   &72.23\% &64.71\% &64.10\%    &65.59\% &58.64\% &73.02\% &80.73\%       &67.89\% &5\\
LLaVA-1.5               &53.85\% &70.16\% &67.40\% &59.93\%   &69.10\% &65.71\% &62.37\%    &62.36\% &58.92\% &70.71\% &81.22\%       &66.32\% &6\\
Qwen-VL                 &54.81\% &66.25\% &62.91\% &60.64\%   &68.30\% &58.85\% &59.44\%    &61.25\% &55.38\% &67.53\% &74.15\%       &63.21\% &7\\
LLaVA                   &46.79\% &63.59\% &65.30\% &64.54\%   &64.29\% &61.10\% &60.77\%    &65.39\% &52.27\% &61.18\% &74.88\%       &62.43\% &8\\
InstructBLIP            &37.82\% &55.36\% &55.43\% &57.09\%   &57.06\% &55.86\% &47.21\%    &59.84\% &45.01\% &54.98\% &56.34\%       &54.29\% &9\\
MiniGPT-v2              &56.73\% &56.44\% &51.74\% &50.00\%   &56.74\% &53.24\% &50.93\%    &53.99\% &43.06\% &58.73\% &66.10\%       &54.18\% &10\\
GLM                     &55.77\% &54.61\% &51.25\% &48.94\%   &54.90\% &55.24\% &47.34\%    &60.95\% &44.62\% &48.48\% &55.61\%       &52.96\% &11\\
Otter                   &35.90\% &54.28\% &51.65\% &51.06\%   &51.04\% &50.62\% &51.20\%    &56.10\% &44.48\% &51.37\% &49.02\%       &50.96\% &12\\
IDEFICS-Instruct        &37.50\% &52.87\% &52.84\% &51.06\%   &52.97\% &50.12\% &48.40\%    &50.96\% &44.62\% &51.09\% &60.73\%       &50.82\% &13\\
MiniGPT-4               &39.42\% &41.31\% &42.67\% &44.33\%   &41.57\% &42.89\% &41.36\%    &47.23\% &32.01\% &41.99\% &46.10\%       &41.93\% &14\\
TinyGPT-V               &21.79\% &24.52\% &22.13\% &28.01\%   &22.71\% &24.69\% &24.34\%    &32.39\% &17.99\% &19.77\% &19.27\%       &23.71\% &15\\
\bottomrule
\end{tabular}}
\end{table*}

\begin{table*}[!t]
\caption{Evaluation results on the \textbf{Aesthetic Empathy} ability.}  \label{AesE}
\renewcommand{\arraystretch}{1.5}
\fontsize{8.5pt}{7pt}\selectfont %
\centering
\setlength{\tabcolsep}{0.9mm}{
\begin{tabular}{l|cccc|ccc|cccc|cc}
\toprule
\multirow{2}*{MLLM}  & \multicolumn{4}{c|}{\textbf{Empathy Dimensions}}   & \multicolumn{3}{c|}{\textbf{Image Sources}}  & \multicolumn{4}{c|}{\textbf{Question Types}}   & {\multirow{2}*{\textbf{Overall}}}& {\multirow{2}*{\textbf{Rank}}}\\
\cmidrule(lr){2-5}\cmidrule(lr){6-8}\cmidrule(lr){9-12}
&\emph{Emotion} &\emph{Interest} &\emph{Uniqueness} &\emph{Vibe} &\emph{NIs} &\emph{AIs} &\emph{AGIs}   &\emph{Yes-No} &\emph{What} &\emph{How} &\emph{Why}  \\
\midrule 
\rowcolor{lightgray!25}
\emph{Random guess}     &40.28\% &37.80\% &36.25\% &42.12\%    &33.10\% &30.62\% &30.67\%    &50.00\% &27.35\% &27.32\% &25.00\%       &34.98\% &/\\
\midrule 
Q-Instruct              &\textbf{68.64\%} &83.86\% &\textbf{75.86\%} &\underline{80.00\%}    &\textbf{76.65\%} &72.19\% &66.62\%    &64.30\% &\textbf{67.42\%} &\textbf{81.57\%} &\underline{86.76\%}      &\textbf{72.68\%} &\textbf{1}\\
Gemini Pro Vision       &\underline{66.87\%} &\textbf{87.50\%} &70.00\% &79.09\%    &70.60\% &\underline{72.35\%} &\textbf{71.53\%}    &67.50\% &64.52\% &72.25\% &\textbf{90.37\%}       &\underline{71.37\%} &\underline{2}\\
ShareGPT4V              &66.48\% &80.65\% &68.97\% &78.72\%    &70.95\% &\textbf{73.69\%} &\underline{67.29\%}    &67.75\% &\underline{65.58\%} &\underline{72.71\%} &83.58\%       &70.75\% &\textbf{3}\\
GPT-4V                  &65.06\% &72.41\% &62.07\% &\textbf{80.15\%}    &\underline{73.87\%} &72.08\% &62.27\%    &\textbf{68.67\%} &64.02\% &70.07\% &84.20\%       &70.16\% &4\\
mPLUG-Owl2              &65.60\% &77.42\% &65.52\% &78.07\%    &71.03\% &71.57\% &66.22\%    &\underline{68.05\%} &64.16\% &70.14\% &83.82\%       &69.89\% &5\\
LLaVA-1.5              &62.49\% &80.65\% &\underline{75.85\%} &78.93\%    &69.26\% &69.58\% &65.43\%    &62.37\% &64.16\% &71.71\% &84.07\%       &68.32\% &6\\
LLaVA                &58.61\% &80.63\% &65.52\% &75.83\%    &67.01\% &66.96\% &58.38\%    &67.95\% &55.95\% &60.14\% &79.66\%       &64.68\% &7\\
Qwen-VL                 &58.67\% &\underline{83.87\%} &72.41\% &73.90\%    &63.88\% &67.08\% &61.57\%    &60.65\% &58.07\% &66.14\% &79.90\%       &64.18\% &8\\
MiniGPT-v2              &52.52\% &58.06\% &44.83\% &58.07\%    &55.86\% &55.85\% &50.27\%    &57.81\% &43.48\% &53.43\% &66.42\%       &54.36\% &9\\
GLM                     &53.13\% &70.97\% &44.83\% &55.29\%    &56.58\% &54.86\% &48.67\%    &60.65\% &41.78\% &50.43\% &64.95\%       &53.96\% &10\\
InstructBLIP            &49.64\% &58.06\% &51.72\% &61.50\%    &55.06\% &55.24\% &48.94\%    &55.88\% &50.99\% &51.43\% &58.33\%       &53.89\% &11\\
Otter                   &48.42\% &70.97\% &51.72\% &63.21\%    &53.05\% &55.74\% &52.39\%    &54.77\% &51.84\% &53.43\% &54.41\%       &53.64\% &12\\
IDEFICS-Instruct        &43.93\% &64.52\% &62.07\% &64.06\%    &50.72\% &53.12\% &49.07\%    &50.20\% &41.08\% &52.43\% &66.42\%       &50.82\% &13\\
MiniGPT-4               &39.78\% &38.71\% &24.14\% &39.04\%    &42.70\% &37.78\% &35.51\%    &50.61\% &31.59\% &31.86\% &38.48\%       &39.35\% &14\\
TinyGPT-V               &30.36\% &29.03\% &31.03\% &35.40\%    &32.50\% &36.03\% &26.99\%    &36.00\% &29.89\% &28.86\% &31.62\%       &32.04\% &15\\

\bottomrule
\end{tabular}}
\end{table*}

\subsection{Benchmark on Aesthetic Assessment}

In the third task of AesBench, the objective is to evaluate the abilities of MLLMs to quantify the aesthetic quality of images. Specifically, in the AesAQA subset, we collect three aesthetic ratings for 2,800 images from aesthetic experts, including \emph{high}, \emph{medium}, and \emph{low}. Then, we feed the quality-related questions to various MLLMs to measure the simple task of \textbf{aesthetic three-grade classification}, as illustrated in Figure \ref{questions}. The input prompt is exemplified as:
\emph{\#User: How is the aesthetic quality of this image?} [Image\_token] \emph{Choose one from the following options: A) High B) Medium C) Low. You should output the correct option without explanation.}
Finally, we leverage the GPT-assisted evaluation strategy to calculate the accuracy rate of the responses generated by MLLMs.

Although the numerical prediction of MLLMs is not the research focus in the community, we still try to explore the abilities of MLLMs to generate \textbf{fine-grained aesthetic scores}. Generally, the outputs of MLLMs are usually verbal text rather than a definite numerical value, and the existing work indicates that the direct output of quality score does not apply to existing MLLMs \cite{Q-Bench}. Therefore, how to collect quantifiable scores from MLLMs remains challenging. Inspired by recent studies on text-based aesthetic assessment \cite{AesCLIP,captioning2}, we introduce a text-to-score approach to convert aesthetic quality descriptions of MLLMs into aesthetic quality scores via a BERT-based regressor \cite{BERT}. Technical details and results are provided in {\color{gray}Supplementary Sections 2 and 3}.

\subsection{Benchmark on Aesthetic Interpretation}
Language interpretation ability is an important feature of MLLMs \cite{LLAVA}. In the last task of AesBench, we examine whether current MLLMs can accurately interpret the reasons for the aesthetic quality. In the AesInter subset, we collect expert interpretations for each of the 2,800 images, as shown in Figure 2. Then, we feed the interpretation-related questions to MLLMs for evaluation. The input prompt is exemplified as:
\#User: \emph{This is an image with [high] aesthetics, please explain the reasons that are related to the aesthetic quality}, where \emph{[high]} is replaced by the actual aesthetic quality grades. The purpose of this is, in the case of known true aesthetic quality, to explore whether current MLLMs can give an aesthetic interpretation that is consistent with aesthetic experts. To evaluate this, we refer to recent works \cite{Judging,Q-Bench}, which have validated the efficacy of using a single-modal GPT as a reliable evaluation tool for pure language tasks. Leveraging this paradigm, this task is converted into a text-only evaluation by matching the MLLM outputs with the expert interpretations under three dimensions: \textbf{Relevance}, \textbf{Precision}, and \textbf{Completeness}, ensuring a comprehensive benchmark on the aesthetic interpretation. The configurations of the GPT-based evaluation are delineated in {\color{gray}Supplementary Section 2}.

 \section{Experiments on AesBench}
\subsection{Experimental Setting}
In this section, we conduct extensive experiments with 15 different MLLMs to evaluate their abilities on image aesthetics perception, including the popular GPT-4V \cite{GPT-4V} and Gemini Pro Vision \cite{Gemini}, as well as 13 state-of-the-art variants with open sources, \emph{i.e.} LLaVA (LLaMA-2-Chat-7B) \cite{LLAVA}, LLaVA-1.5 (Vicuna-v1.5-7B) \cite{LLaVA-1.5}, ShareGPT4V (Vicuna-v1.5-7B) \cite{ShareGPT4V}, Q-Instruct (Vicuna-v1.5-7B) \cite{Q-Instruct}, mPLUG-Owl2 (LLaMA-2-7B) \cite{mPLUG-Owl2}, InstructBLIP (Vicuna-7B) \cite{InstructBLIP}, MiniGPT-4 (Vicuna-7B) \cite{MiniGPT-4}, MiniGPT-v2 (LLaMA-2-Chat-7B) \cite{MiniGPT-v2}, IDEFICS-Instruct (LLaMA-7B) \cite{IDEFICS}, GLM (ChatGLM-6B) \cite{GLM} Otter (MPT-7B) \cite{Otter}, TinyGPT-V (Phi-2) \cite{TinyGPT-V} and Qwen-VL (QWen-7B) \cite{Qwen-VL}. To ensure integrity and fairness, these MLLMs are evaluated using the original model weights as released by the authors without any dataset-specific tuning. It is worth noting that GPT-4V and Gemini Pro Vision have not been made available in the public domain as open-source entities. To facilitate experimental analysis, their capabilities are gauged through responses obtained via APIs. More experimental details and analyses are provided in {\color{gray}Supplementary Section 3}.

\subsection{Evaluation on Aesthetic Perception}
To comprehensively evaluate the aesthetic perception abilities of current MLLMs, we measure the accuracy of responses produced by these MLLMs across various perceptual dimensions, image sources, and question types using the AesPQA subset. The results are listed in Table \ref{AesP}. Notably, Q-Instruct achieves superior overall performance in aesthetic perception. For different perceptual dimensions, GPT-4V performs best in terms of  \emph{Technical Quality} and  \emph{Color and Light}, while Q-Instruct outperforms other MLLMs in terms of  \emph{Composition} and  \emph{Content}. We also notice that most MLLMs perform best on natural images, followed by artistic images, and worst on artificial intelligence-generated images, which potentially correlates with the data composition in their training sets. In conclusion, AesBench can effectively discriminate among the aesthetic perception capabilities of MLLMs.

\subsection{Evaluation on Aesthetic Empathy}
For evaluating the aesthetic empathy capabilities of current MLLMs, we test their accuracy in generating responses across various empathy dimensions, image sources, and question types, utilizing the AesEQA subset. Table \ref{AesE} summarizes the experimental results. The results indicate that among the evaluated MLLMs, Q-Instruct exhibits the highest overall performance in aesthetic empathy, while Gemini Pro Vision achieves the second-best overall performance. For different empathy dimensions, Q-Instruct outperforms other MLLMs in terms of \emph{Emotion} and  \emph{Uniqueness}; Gemini Pro Vision reaches the best accuracy on \emph{Interest}, and GPT-4V performs best on \emph{Vibe}. TinyGPT performs worst overall, even lower than the accuracy of random guesses. In summary, we comprehensively evaluate the strengths and weaknesses of current MLLMs in aesthetic empathy.

\begin{table}[!t]
\caption{Evaluation results on the \textbf{Aesthetic Assessment} ability.}  \label{AesA}
\renewcommand{\arraystretch}{1.6}
\fontsize{8.5pt}{7pt}\selectfont %
\centering
\setlength{\tabcolsep}{1.3mm}{
\begin{tabular}{l|ccc|ccc|cc}
\toprule

MLLM     &\emph{NIs} &\emph{AIs} &\emph{AGIs }  &\textbf{Overall} &\textbf{Rank}\\
      \midrule 
\rowcolor{lightgray!25}
\emph{Random guess}     &33.33\% &33.33\% &33.33\%    &33.33\%    &{/}   \\
   \midrule 
Q-Instruct              &\textbf{62.20\%} &\textbf{49.75\%} &40.69\%    &\textbf{52.86\%}    &\textbf{1}  \\
GPT-4V                  &\underline{59.98\%} &46.92\% &40.59\%    &\underline{50.86\%}    &\underline{2}  \\
mPLUG-Owl2              &57.78\% &\underline{49.50\%} &\underline{40.83\%}    &50.57\%    &3   \\
Gemini Pro Vision       &54.17\% &48.39\% &\textbf{42.20\%}    &49.38\%    &4  \\
ShareGPT4V              &54.65\% &48.38\% &35.90\%    &47.82\%    &5   \\
InstructBLIP            &52.73\% &47.88\% &34.84\%    &46.54\%    &6   \\
Qwen-VL                 &54.25\% &39.28\% &40.43\%    &46.25\%    &7   \\
LLaVA                &51.69\% &48.00\% &34.31\%    &45.96\%    &8  \\
LLaVA-1.5              &50.08\% &48.13\% &34.97\%    &45.46\%    &9  \\
IDEFICS-Instruct                 &50.00\% &47.76\% &33.78\%    &45.00\%    &10   \\
Otter                   &49.20\% &48.25\% &34.04\%    &44.86\%    &11   \\
TinyGPT-V               &44.06\% &41.65\% &44.81\%    &43.57\%    &12   \\
MiniGPT-4               &41.65\% &36.28\% &35.90\%    &38.57\%    &13   \\
GLM                     &38.92\% &37.78\% &35.90\%    &37.79\%    &14   \\
MiniGPT-v2              &27.05\% &31.92\% &36.97\%    &31.11\%    &15   \\

\bottomrule
\end{tabular}}
\end{table}

\begin{table}[!t]
\caption{Evaluation results on the \textbf{Aesthetic Interpretation} ability.} \label{AesI}
\renewcommand{\arraystretch}{1.6}
\fontsize{8.5pt}{7pt}\selectfont %
\centering
\setlength{\tabcolsep}{1.9mm}{
\begin{tabular}{l|ccc|cc}
\toprule
MLLM  &\emph{Rele.} &\emph{Prec.} &\emph{Comp.} &\textbf{Overall} &\textbf{Rank}\\
\midrule 
GPT-4V                    &1.385 &\textbf{1.151} &\textbf{1.366} &\textbf{1.301}    &\textbf{1}\\
ShareGPT4V                &\textbf{1.440} &\underline{1.117} &\underline{1.331} &\underline{1.296}    &\underline{2}   \\
Gemini Pro Vision         &\underline{1.416} &1.087 &1.164 &1.222    &3   \\
Qwen-VL                   &1.393 &1.006 &1.175 &1.192    &4\\
mPLUG-Owl2                &1.402 &1.016 &1.130 &1.182    &5   \\
IDEFICS-Instruct          &1.406 &1.007 &1.126 &1.180    &6   \\
LLaVA-1.5                 &1.397 &0.953 &1.120 &1.157    &7   \\
InstructBLIP              &1.372 &0.863 &1.144 &1.126    &8   \\
LLaVA                     &1.374 &0.918 &1.084 &1.125    &9   \\
Otter                     &1.242 &0.848 &0.989 &1.027    &10   \\
Q-Instruct                &1.222 &0.939 &0.898 &1.020    &11   \\
MiniGPT-v2                &1.191 &0.868 &0.948 &1.003    &12   \\
MiniGPT-4                 &1.158 &0.823 &1.016 &0.999    &13   \\
GLM                       &1.122 &0.729 &0.944 &0.932    &14   \\
TinyGPT-V                 &0.871 &0.511 &0.720 &0.701    &15   \\

\bottomrule
\end{tabular}}
\end{table}

\subsection{Evaluation on Aesthetic Assessment}
To benchmark the aesthetic assessment capabilities, we evaluate the performance of existing MLLMs on the AesAQA subset. The experimental results are reported in Table \ref{AesA}. The results indicate that Q-Instruct achieves superior performance in overall accuracy for aesthetic classification. However, it is notable that the highest recorded prediction accuracy is only 52.86\%, underscoring the limitations of even the most advanced MLLMs in accurately assessing aesthetic quality. Additionally, GPT-4V emerges as the model with the second-highest prediction accuracy, at 50.86\%. An interesting observation from this table is the varying performance of MLLMs across different image types: the models are most effective with natural images, moderately effective with artistic images, and least effective with images generated by artificial intelligence. Furthermore, we also try to explore the abilities of MLLMs to generate fine-grained aesthetic scores, and results are provided in {\color{gray}Supplementary Section 3}. These experiments demonstrate that the aesthetic assessment abilities of MLLMs still need to be further strengthened in the future.

\subsection{Evaluation on Aesthetic Interpretation}
We benchmark the interpretation abilities of current MLLMs in Table \ref{AesI}, where the values range from [0, 2], and the higher the value, the better the performance. From the table, GPT-4V is superior to other MLLMs in overall performance, while ShareGPT obtains the second-best overall performance. For the three different dimensions, we can find that almost all models suffer from \emph{unsatisfactory Precision}, significantly less than the other dimensions (\emph{Relevance} and \emph{Completeness}). The significantly lower precision highlights the strong \textbf{hallucination} in the context of aesthetic interpretation, which underscores the crucial research for enhancing the aesthetic interpretation abilities of MLLMs, especially on language precision.

 \section{Conclusion}

In this work, we introduce AesBench, an expert benchmark to comprehensively evaluate the aesthetic perception capacities of the current MLLMs. AesBench comprises two main components. First, the Expert-labeled Aesthetics Perception Database (EAPD), a high-quality dataset that serves as a gold standard for evaluating MLLMs in image aesthetics perception. Second, a four-dimensional evaluation framework is proposed, which consists of a set of comprehensive criteria across four key dimensions: Perception (AesP), Empathy (AesE), Assessment (AesA), and Interpretation (AesI). Experimental evaluations on AesBench demonstrate that most MLLMs are still weak in aesthetic perception capabilities. This also highlights that future development of MLLMs needs to consider more aesthetic dimensions.

\bibliographystyle{named}
\bibliography{ijcai24}

\begin{thebibliography}{}

\bibitem[\protect\citeauthoryear{Achlioptas \bgroup \em et al.\egroup }{2021}]{ArtEmis}
Panos Achlioptas, Maks Ovsjanikov, Kilichbek Haydarov, Mohamed Elhoseiny, and Leonidas Guibas.
\newblock Artemis: Affective language for visual art.
\newblock In {\em Proc. IEEE Conf. Comput. Vis. Pattern Recognit.}, pages 11564--11574, 2021.

\bibitem[\protect\citeauthoryear{Antol \bgroup \em et al.\egroup }{2015}]{VQA}
Stanislaw Antol, Aishwarya Agrawal, Jiasen Lu, Margaret Mitchell, Dhruv Batra, C.~Lawrence Zitnick, and Devi Parikh.
\newblock {VQA}: Visual question answering.
\newblock In {\em Proc. IEEE Int. Conf. on Comput. Vis.}, December 2015.

\bibitem[\protect\citeauthoryear{Bai \bgroup \em et al.\egroup }{2023}]{Qwen-VL}
Jinze Bai, Shuai Bai, Shusheng Yang, Shijie Wang, Sinan Tan, Peng Wang, Junyang Lin, Chang Zhou, and Jingren Zhou.
\newblock {Qwen-VL}: A versatile vision-language model for understanding, localization, text reading, and beyond.
\newblock {\em arXiv preprint arXiv:2308.12966}, 2023.

\bibitem[\protect\citeauthoryear{Cai \bgroup \em et al.\egroup }{2023}]{BenchLMM}
Rizhao Cai, Zirui Song, Dayan Guan, Zhenhao Chen, Xing Luo, Chenyu Yi, and Alex Kot.
\newblock {BenchLMM}: Benchmarking cross-style visual capability of large multimodal models.
\newblock {\em arXiv preprint arXiv:2312.02896}, 2023.

\bibitem[\protect\citeauthoryear{Chen \bgroup \em et al.\egroup }{2023a}]{MiniGPT-v2}
Jun Chen, Deyao Zhu, Xiaoqian Shen, Xiang Li, Zechu Liu, Pengchuan Zhang, Raghuraman Krishnamoorthi, Vikas Chandra, Yunyang Xiong, and Mohamed Elhoseiny.
\newblock {MiniGPT-v2}: large language model as a unified interface for vision-language multi-task learning.
\newblock {\em arXiv preprint arXiv:2310.09478}, 2023.

\bibitem[\protect\citeauthoryear{Chen \bgroup \em et al.\egroup }{2023b}]{ShareGPT4V}
Lin Chen, Jisong Li, Xiaoyi Dong, Pan Zhang, Conghui He, Jiaqi Wang, Feng Zhao, and Dahua Lin.
\newblock {ShareGPT4V}: Improving large multi-modal models with better captions.
\newblock {\em arXiv preprint arXiv:2311.12793}, 2023.

\bibitem[\protect\citeauthoryear{Chen \bgroup \em et al.\egroup }{2023c}]{EgoPlan-Bench}
Yi~Chen, Yuying Ge, Yixiao Ge, Mingyu Ding, Bohao Li, Rui Wang, Ruifeng Xu, Ying Shan, and Xihui Liu.
\newblock {EgoPlan-Bench}: Benchmarking egocentric embodied planning with multimodal large language models.
\newblock {\em arXiv preprint arXiv:2312.06722}, 2023.

\bibitem[\protect\citeauthoryear{Dai \bgroup \em et al.\egroup }{2023}]{InstructBLIP}
Wenliang Dai, Junnan Li, Dongxu Li, Anthony Meng~Huat Tiong, Junqi Zhao, Weisheng Wang, Boyang Li, Pascale Fung, and Steven Hoi.
\newblock {InstructBLIP}: Towards general-purpose vision-language models with instruction tuning.
\newblock {\em arXiv preprint arXiv:2305.06500}, 2023.

\bibitem[\protect\citeauthoryear{Devlin \bgroup \em et al.\egroup }{2019}]{BERT}
Jacob Devlin, Ming-Wei Chang, Kenton Lee, and Kristina Toutanova.
\newblock {BERT}: Pre-training of deep bidirectional transformers for language understanding.
\newblock {\em arXiv preprint arXiv:1810.04805}, 2019.

\bibitem[\protect\citeauthoryear{Du \bgroup \em et al.\egroup }{2022}]{GLM}
Zhengxiao Du, Yujie Qian, Xiao Liu, Ming Ding, Jiezhong Qiu, Zhilin Yang, and Jie Tang.
\newblock {GLM}: General language model pretraining with autoregressive blank infilling.
\newblock {\em arXiv preprint arXiv:2103.10360}, 2022.

\bibitem[\protect\citeauthoryear{Fang \bgroup \em et al.\egroup }{2020}]{SPAQ}
Yuming Fang, Hanwei Zhu, Yan Zeng, Kede Ma, and Zhou Wang.
\newblock Perceptual quality assessment of smartphone photography.
\newblock In {\em Proc. IEEE Conf. Comput. Vis. Pattern Recognit.}, pages 3674--3683, June 2020.

\bibitem[\protect\citeauthoryear{Feng \bgroup \em et al.\egroup }{2023}]{EmP}
Tinglei Feng, Jiaxuan Liu, and Jufeng Yang.
\newblock Probing sentiment-oriented pretraining inspired by human sentiment perception mechanism.
\newblock In {\em Proc. IEEE Conf. Comput. Vis. Pattern Recognit.}, pages 2850--2860, Jun. 2023.

\bibitem[\protect\citeauthoryear{Google}{2023}]{Gemini}
Google.
\newblock Build with gemini, 2023.

\bibitem[\protect\citeauthoryear{He \bgroup \em et al.\egroup }{2022}]{TAD66K}
Shuai He, Yongchang Zhang, Rui Xie, Dongxiang Jiang, and Anlong Ming.
\newblock Rethinking image aesthetics assessment: Models, datasets and benchmarks.
\newblock {\em Proc. Int. Joint Conf. Artif. Intell.}, Jul. 2022.

\bibitem[\protect\citeauthoryear{Hossain \bgroup \em et al.\egroup }{2019}]{Captioning}
MD~Zakir Hossain, Ferdous Sohel, Mohd~Fairuz Shiratuddin, and Hamid Laga.
\newblock A comprehensive survey of deep learning for image captioning.
\newblock {\em ACM Computing Surveys (CsUR)}, 51(6):1--36, 2019.

\bibitem[\protect\citeauthoryear{Hosu \bgroup \em et al.\egroup }{2020}]{KonIQ-10K}
Vlad Hosu, Hanhe Lin, Tamas Sziranyi, and Dietmar Saupe.
\newblock {KonIQ-10k}: An ecologically valid database for deep learning of blind image quality assessment.
\newblock {\em IEEE Trans. Image Process.}, 29:4041--4056, Jan. 2020.

\bibitem[\protect\citeauthoryear{Huang \bgroup \em et al.\egroup }{2020}]{DSS}
Yipo Huang, Leida Li, Hancheng Zhu, and Bo~Hu.
\newblock Blind quality index of depth images based on structural statistics for view synthesis.
\newblock {\em IEEE Signal Process. Lett.}, 27:685--689, Apr. 2020.

\bibitem[\protect\citeauthoryear{Huang \bgroup \em et al.\egroup }{2023}]{SARQUE}
Yipo Huang, Leida Li, Yuzhe Yang, Yaqian Li, and Yandong Guo.
\newblock Explainable and generalizable blind image quality assessment via semantic attribute reasoning.
\newblock {\em IEEE Trans. Multimedia}, 25:7672--7685, 2023.

\bibitem[\protect\citeauthoryear{Huggingface}{2023}]{IDEFICS}
Huggingface.
\newblock Introducing idefics: An open reproduction of state-of-the-art visual language model, 2023.

\bibitem[\protect\citeauthoryear{Kong \bgroup \em et al.\egroup }{2016}]{AADB}
S.~Kong, X.~Shen, Z.~Lin, R.~Mech, and C.~Fowlkes.
\newblock Photo aesthetics ranking network with attributes and content adaptation.
\newblock In {\em Proc. Eur. Conf. Comput. Vis.}, pages 662--679, Sep. 2016.

\bibitem[\protect\citeauthoryear{Kruk \bgroup \em et al.\egroup }{2023}]{Impressions}
Julia Kruk, Caleb Ziems, and Diyi Yang.
\newblock Impressions: Understanding visual semiotics and aesthetic impact.
\newblock {\em arXiv preprint arXiv:2310.17887}, 2023.

\bibitem[\protect\citeauthoryear{Lai \bgroup \em et al.\egroup }{2023}]{Segmentation}
Xin Lai, Zhuotao Tian, Yukang Chen, Yanwei Li, Yuhui Yuan, Shu Liu, and Jiaya Jia.
\newblock {LISA}: Reasoning segmentation via large language model.
\newblock {\em arXiv preprint arXiv:2308.00692}, 2023.

\bibitem[\protect\citeauthoryear{Li \bgroup \em et al.\egroup }{2018}]{CAD}
Bo~Li, Caiming Xiong, Tianfu Wu, Yu~Zhou, Lun Zhang, and Rufeng Chu.
\newblock Neural abstract style transfer for chinese traditional painting.
\newblock {\em arXiv preprint arXiv:1812.03264}, 2018.

\bibitem[\protect\citeauthoryear{Li \bgroup \em et al.\egroup }{2023a}]{Otter}
Bo~Li, Yuanhan Zhang, Liangyu Chen, Jinghao Wang, Jingkang Yang, and Ziwei Liu.
\newblock Otter: A multi-modal model with in-context instruction tuning.
\newblock {\em arXiv preprint arXiv:2305.03726}, 2023.

\bibitem[\protect\citeauthoryear{Li \bgroup \em et al.\egroup }{2023b}]{AGIQA-3K}
Chunyi Li, Zicheng Zhang, Haoning Wu, Wei Sun, Xiongkuo Min, Xiaohong Liu, Guangtao Zhai, and Weisi Lin.
\newblock {AGIQA-3K}: An open database for ai-generated image quality assessment.
\newblock {\em arXiv preprint arXiv:2306.04717}, 2023.

\bibitem[\protect\citeauthoryear{Li \bgroup \em et al.\egroup }{2023c}]{TAVAR}
Leida Li, Yipo Huang, Jinjian Wu, Yuzhe Yang, Yaqian Li, Yandong Guo, and Guangming Shi.
\newblock Theme-aware visual attribute reasoning for image aesthetics assessment.
\newblock {\em IEEE Trans. Circuits and Syst. Video Technol.}, 2023.

\bibitem[\protect\citeauthoryear{Liu \bgroup \em et al.\egroup }{2023a}]{LLaVA-1.5}
Haotian Liu, Chunyuan Li, Yuheng Li, and Yong~Jae Lee.
\newblock Improved baselines with visual instruction tuning.
\newblock {\em arXiv preprint arXiv:2310.03744}, 2023.

\bibitem[\protect\citeauthoryear{Liu \bgroup \em et al.\egroup }{2023b}]{LLAVA}
Haotian Liu, Chunyuan Li, Qingyang Wu, and Yong~Jae Lee.
\newblock Visual instruction tuning.
\newblock {\em arXiv preprint arXiv:2304.08485}, 2023.

\bibitem[\protect\citeauthoryear{Lu \bgroup \em et al.\egroup }{2021}]{MM3}
Peng Lu, Hao Zhang, Xujun Peng, and Xiaofu Jin.
\newblock Learning the relation between interested objects and aesthetic region for image cropping.
\newblock {\em IEEE Trans. Multimedia}, 23:3618--3630, Oct. 2021.

\bibitem[\protect\citeauthoryear{OpenAI}{2023}]{chatgpt}
OpenAI.
\newblock Introducing chatgpt, 2023.

\bibitem[\protect\citeauthoryear{Radford \bgroup \em et al.\egroup }{2021}]{CLIP}
Alec Radford, Jong~Wook Kim, Chris Hallacy, Aditya Ramesh, Gabriel Goh, Sandhini Agarwal, Girish Sastry, Amanda Askell, Pamela Mishkin, Jack Clark, Gretchen Krueger, and Ilya Sutskever.
\newblock Learning transferable visual models from natural language supervision.
\newblock {\em arXiv preprint arXiv:2103.00020}, 2021.

\bibitem[\protect\citeauthoryear{Sheikh \bgroup \em et al.\egroup }{2006}]{Evaluation}
H.R. Sheikh, M.F. Sabir, and A.C. Bovik.
\newblock A statistical evaluation of recent full reference image quality assessment algorithms.
\newblock {\em IEEE Trans. Image Process.}, 15(11):3440--3451, Nov. 2006.

\bibitem[\protect\citeauthoryear{Sheng \bgroup \em et al.\egroup }{2023}]{AesCLIP}
Xiangfei Sheng, Leida Li, Pengfei Chen, Jinjian Wu, Weisheng Dong, Yuzhe Yang, Liwu Xu, Yaqian Li, and Guangming Shi.
\newblock Aesclip: Multi-attribute contrastive learning for image aesthetics assessment.
\newblock In {\em Proc. ACM Int. Conf. Multimedia}, page 1117–1126, Oct. 2023.

\bibitem[\protect\citeauthoryear{Touvron \bgroup \em et al.\egroup }{2023}]{LLaMA}
Hugo Touvron, Thibaut Lavril, Gautier Izacard, Xavier Martinet, Marie-Anne Lachaux, Timothée Lacroix, Baptiste Rozière, Naman Goyal, Eric Hambro, Faisal Azhar, Aurelien Rodriguez, Armand Joulin, Edouard Grave, and Guillaume Lample.
\newblock {LLaMA}: Open and efficient foundation language models.
\newblock {\em arXiv preprint arXiv:2302.13971}, 2023.

\bibitem[\protect\citeauthoryear{Unsplash}{2023}]{LITE}
Unsplash.
\newblock Access the world’s largest open library dataset, 2023.

\bibitem[\protect\citeauthoryear{Wang \bgroup \em et al.\egroup }{2023}]{DiffusionDB}
Zijie~J. Wang, Evan Montoya, David Munechika, Haoyang Yang, Benjamin Hoover, and Duen~Horng Chau.
\newblock {DiffusionDB}: A large-scale prompt gallery dataset for text-to-image generative models.
\newblock {\em arXiv preprint arXiv:2210.14896}, 2023.

\bibitem[\protect\citeauthoryear{Wu \bgroup \em et al.\egroup }{2023a}]{Q-Instruct}
Haoning Wu, Zicheng Zhang, Erli Zhang, Chaofeng Chen, Liang Liao, Annan Wang, Kaixin Xu, Chunyi Li, Jingwen Hou, Guangtao Zhai, Geng Xue, Wenxiu Sun, Qiong Yan, and Weisi Lin.
\newblock {Q-Instruct}: Improving low-level visual abilities for multi-modality foundation models.
\newblock {\em arXiv preprint arXiv:2311.06783}, 2023.

\bibitem[\protect\citeauthoryear{Wu \bgroup \em et al.\egroup }{2023b}]{Q-Align}
Haoning Wu, Zicheng Zhang, Weixia Zhang, Chaofeng Chen, Liang Liao, Chunyi Li, Yixuan Gao, Annan Wang, Erli Zhang, Wenxiu Sun, Qiong Yan, Xiongkuo Min, Guangtao Zhai, and Weisi Lin.
\newblock Q-align: Teaching lmms for visual scoring via discrete text-defined levels.
\newblock {\em arXiv preprint arXiv:2312.17090}, 2023.

\bibitem[\protect\citeauthoryear{Wu \bgroup \em et al.\egroup }{2024}]{Q-Bench}
Haoning Wu, Zicheng Zhang, Erli Zhang, Chaofeng Chen, Liang Liao, Annan Wang, Chunyi Li, Wenxiu Sun, Qiong Yan, Guangtao Zhai, and Weisi Lin.
\newblock {Q-Bench}: A benchmark for general-purpose foundation models on low-level vision.
\newblock {\em arXiv preprint arXiv:2309.14181}, 2024.

\bibitem[\protect\citeauthoryear{Yang \bgroup \em et al.\egroup }{2022}]{PARA}
Y.~Yang, L.~Xu, L.~Li, N.~Qie, Y.~Li, P.~Zhang, and Y.~Guo.
\newblock Personalized image aesthetics assessment with rich attributes.
\newblock In {\em Proc. IEEE Conf. Comput. Vis. Pattern Recog.}, pages 19861--19869, Jun. 2022.

\bibitem[\protect\citeauthoryear{Yang \bgroup \em et al.\egroup }{2023}]{GPT-4V}
Zhengyuan Yang, Linjie Li, Kevin Lin, Jianfeng Wang, Chung-Ching Lin, Zicheng Liu, and Lijuan Wang.
\newblock The dawn of lmms: Preliminary explorations with gpt-4v(ision).
\newblock {\em arXiv preprint arXiv:2309.17421}, 2023.

\bibitem[\protect\citeauthoryear{Ye \bgroup \em et al.\egroup }{2023}]{mPLUG-Owl2}
Qinghao Ye, Haiyang Xu, Jiabo Ye, Ming Yan, Anwen Hu, Haowei Liu, Qi~Qian, Ji~Zhang, Fei Huang, and Jingren Zhou.
\newblock {mPLUG-Owl2}: Revolutionizing multi-modal large language model with modality collaboration.
\newblock {\em arXiv preprint arXiv:2311.04257}, 2023.

\bibitem[\protect\citeauthoryear{Yi \bgroup \em et al.\egroup }{2023}]{BAID}
Ran Yi, Haoyuan Tian, Zhihao Gu, Yu-Kun Lai, and Paul~L. Rosin.
\newblock Towards artistic image aesthetics assessment: a large-scale dataset and a new method.
\newblock In {\em Proc. IEEE Conf. Comput. Vis. Pattern Recognit.}, pages 22388--22397, 2023.

\bibitem[\protect\citeauthoryear{Yuan \bgroup \em et al.\egroup }{2023}]{TinyGPT-V}
Zhengqing Yuan, Zhaoxu Li, and Lichao Sun.
\newblock {TinyGPT-V}: Efficient multimodal large language model via small backbones.
\newblock {\em arXiv preprint arXiv:2312.16862}, 2023.

\bibitem[\protect\citeauthoryear{Zhang \bgroup \em et al.\egroup }{2023a}]{AGIQA-1K}
Zicheng Zhang, Chunyi Li, Wei Sun, Xiaohong Liu, Xiongkuo Min, and Guangtao Zhai.
\newblock A perceptual quality assessment exploration for aigc images.
\newblock {\em arXiv preprint arXiv:2303.12618}, 2023.

\bibitem[\protect\citeauthoryear{Zhang \bgroup \em et al.\egroup }{2023b}]{Q-Boost}
Zicheng Zhang, Haoning Wu, Zhongpeng Ji, Chunyi Li, Erli Zhang, Wei Sun, Xiaohong Liu, Xiongkuo Min, Fengyu Sun, Shangling Jui, Weisi Lin, and Guangtao Zhai.
\newblock Q-boost: On visual quality assessment ability of low-level multi-modality foundation models.
\newblock {\em arXiv preprint arXiv:2312.15300}, 2023.

\bibitem[\protect\citeauthoryear{Zheng \bgroup \em et al.\egroup }{2023}]{Judging}
Lianmin Zheng, Wei-Lin Chiang, Ying Sheng, Siyuan Zhuang, Zhanghao Wu, Yonghao Zhuang, Zi~Lin, Zhuohan Li, Dacheng Li, Eric~P. Xing, Hao Zhang, Joseph~E. Gonzalez, and Ion Stoica.
\newblock Judging llm-as-a-judge with mt-bench and chatbot arena.
\newblock {\em arXiv preprint arXiv:2306.05685}, 2023.

\bibitem[\protect\citeauthoryear{Zhong \bgroup \em et al.\egroup }{2023}]{captioning2}
Zhipeng Zhong, Fei Zhou, and Guoping Qiu.
\newblock Aesthetically relevant image captioning.
\newblock In {\em Proc. AAAI}, volume~37, pages 3733--3741, 2023.

\bibitem[\protect\citeauthoryear{Zhu \bgroup \em et al.\egroup }{2023}]{MiniGPT-4}
Deyao Zhu, Jun Chen, Xiaoqian Shen, Xiang Li, and Mohamed Elhoseiny.
\newblock {MiniGPT-4}: Enhancing vision-language understanding with advanced large language models.
\newblock {\em arXiv preprint arXiv:2304.10592}, 2023.

\end{thebibliography}

\end{document}